\documentclass[letterpaper, 10 pt, conference]{ieeeconf}  

\IEEEoverridecommandlockouts                              

\usepackage{amsmath, amsfonts}
\usepackage{xcolor}
\usepackage{float}
\usepackage{graphicx}
\usepackage{wrapfig}

\newcommand{\bx}{\pmb{x}}

\newcommand{\ba}{\pmb{a}}
\newcommand{\bk}{\pmb{k}}
\newcommand{\bs}{\pmb{s}}

\newcommand{\bth}{\pmb{\theta}}

\newcommand{\btau}{\pmb{\tau}}

\newcommand{\bA}{\pmb{A}}
\newcommand{\bS}{\pmb{S}}

\newcommand{\R}{\mathbb{R}}
\newcommand{\rar}{\rightarrow}

\title{\LARGE \bf
A Bayesian Treatment of Real-to-Sim for \\ Deformable Object Manipulation
}

\author{Rika Antonova$^{*1}$, Jingyun Yang$^{2}$, Priya Sundaresan$^{1}$, Dieter Fox$^{3,5}$, Fabio Ramos$^{4,5}$, Jeannette Bohg$^{1}$
\thanks{$^*$Supported by the National Science Foundation grant No.2030859 to the Computing Research Association for the CIFellows Project.}
\thanks{$^{1}$Department of Computer Science, Stanford University, Stanford, CA 94305, USA
        {\tt\scriptsize \{rika.antonova, \ priyasun, \ bohg\}@stanford.edu}}%
\thanks{$^{2}$Machine Learning Department, Carnegie Mellon University, Pittsburgh, PA 15213, USA
        {\tt\scriptsize jingyuny@andrew.cmu.edu}}%
\thanks{$^{3}$NVIDIA, Seattle, WA 98105, USA}%
\thanks{$^{4}$School of Computer Science, The University of Sydney, Darlington, NSW, 2008, Australia
        {\tt\scriptsize fabio.ramos@sydney.edu.au}}%
\thanks{$^{5}$Paul G. Allen School of Computer Science and Engineering, University of Washington, Seattle, WA 98195, USA
{\tt\scriptsize fox@cs.washington.edu}}%
}

\begin{document}

\maketitle
\thispagestyle{empty}
\pagestyle{empty}

\begin{abstract}
Deformable object manipulation remains a challenging task in robotics research. Conventional techniques for parameter inference and state estimation typically rely on a precise definition of the state space and its dynamics. While this is appropriate for rigid objects and robot states, it is challenging to define the state space of a deformable object and how it evolves in time. In this work, we pose the problem of inferring physical parameters of deformable objects as a probabilistic inference task defined with a simulator. We propose a novel methodology for extracting state information from image sequences via a technique to represent the state of a deformable object as a distribution embedding. This allows to incorporate noisy state observations directly into modern Bayesian simulation-based inference tools in a principled manner. Our experiments confirm that we can  estimate posterior distributions of physical properties, such as elasticity, friction and scale of highly deformable objects, such as cloth and ropes. Overall, our method addresses the real-to-sim problem probabilistically and helps to better represent the evolution of the state of deformable objects.
\end{abstract}

\section{Introduction}

The problem of simulation parameter inference received a considerable amount of attention in the broad scientific community: \cite{cranmer2020frontier} provides a recent survey. The `real-to-sim' term is commonly used to refer to parameter inference for large-scale models and general-purpose simulators, recent examples include~\cite{prakash2021self, chang2020sim2real2sim}. More broadly, `real-to-sim' can refer to any problem formulations and methods that help identify and bridge the gap between reality and simulation, e.g.~\cite{zhang2019vr, liu2021real}.
In robotics, this problem has been addressed extensively for the case of rigid objects~\cite{chebotar2019closing, ramos2019bayessim, barcelos2020disco, mehta2020active, muratore2021neural}.
Many of these methods rely on using canonical representations of the state and assume reliable state estimation for obtaining low-dimensional representation of objects in the scene. Deformable objects present unique challenges for defining or learning low-dimensional representations.
Focusing on the relevant parts of their state requires interpreting high-dimensional data, such as images or point clouds. However, computer vision methods that succeed in learning robust low-dimensional representations of rigid objects can fail to learn on the more complex data patterns that the highly deformable objects present.
Even when an acceptable representation is available, the trajectories obtained when manipulating deformable objects still contain complex patterns reflecting the dynamics of the deformables.
Methods that succeed using trajectories containing only rigid objects can fail to interpret the more challenging patterns in the data with deformables, even when the state representation is low-dimensional (our experiments show such examples).

Given these challenges, we advocate the Bayesian treatment and cast the `real-to-sim' problem as simulation-based inference. Bayesian methods can leverage simulation as a source of prior knowledge in a principled way, and can conduct inference from real observations in a data-efficient manner. In this work, we consider likelihood-free inference techniques introduced by~\cite{ramos2019bayessim} that can infer flexible multimodal posteriors, and hence are well-equipped to cope with the challenge of interpreting complex and noisy data patterns in a principled manner.
We investigate how these inference methods behave when the observed deformable object state is noisy and approximate. Keypoint-based representations are promising for deformable objects because of their representational flexibility, ability to quickly obtain a low-dimensional state of an object, and the potential to train from on a small set of observations. For example, we are able to train an unsupervised approach from ~\cite{kulkarni2019unsupervised} with data from only 1 minute of hardware data and 100 simulated trajectories. However, keypoints are not guaranteed to be consistent across frames and tend to appear on different parts of an object. Supervised keypoint extraction techniques can learn to track a certain set of locations reliably, as we show in our experiments with the model from~\cite{sundaresan2020untangling}. However, the keypoints can still permute between these locations even in adjacent frames. Under these conditions, we show that existing inference methods have significant difficulties in real-to-sim experiments, where the real data is obtained from real images of a robot manipulating deformable objects.

Our contribution is a formulation of the problem of real-to-sim for deformable object manipulation as simulation-based inference over a distributional representation of the state.
Specifically, we interpret keypoints on objects extracted from images as samples from an underlying state distribution. We then embed this distribution in a {\em reproducing kernel Hilbert space\/} (RKHS) via the kernel mean embedding operator~\cite{song2013kernel}.
This yields a representation that is permutation-invariant, addressing the case when the keypoints are permuted. Furthermore, this distributional interpretation allows us to avoid the need for keypoints to consistently track exactly the same locations on the object. More generally, this also opens possibilities for using any probabilistic output of vision-based modules, including nonparametric particle-based methods, since RKHS mean embedding can be easily applied in these cases. We call the resulting method \textit{BayesSim-RKHS}.

To show that our approach can successfully handle parameter inference for deformable objects we include experiments on 3 different scenarios on hardware with real deformables: (i) wiping a table surface with cloth; (ii) winding a rope onto a spool; (iii) spreading a highly deformable piece of cloth by flinging it in the air, then dragging it over the table surface. In all these scenarios, \textit{BayesSim-RKHS} significantly outperforms existing BayesSim variants. We demonstrate that this advantage is not tied to a specific keypoint extraction method. We show results that use a recent supervised keypoint method developed specifically for deformables~\cite{sundaresan2020untangling}, and a general unsupervised method that was developed with rigid objects as the primary use case~\cite{kulkarni2019unsupervised}.

\section{Background}
\label{sec:background}

\subsection{Simulating, Representing and Manipulating Deformables}

There are promising results for learning to perceive and manipulate deformable objects, as surveyed in~\cite{herguedas2019survey, arriola2020modeling, yin2021modeling}. However, works in this field usually construct a small self-contained scenario, and do not offer a way to align general-purpose simulators with reality. This is due to the significant challenges in simulation, perception and control of highly deformable objects even when considering only one task or scenario. Typical examples include: manipulating an elastic loop~\cite{yoshida2015simulation}; hanging a piece of cloth~\cite{matas2018sim}; assisting to put on a hat~\cite{klee2015personalized}, a shirt~\cite{clegg2018learning}, a gown~\cite{kapusta2019personalized}, a sleeveless jacket~\cite{shen2021provably}. Setting up these scenarios on hardware requires significant effort. Hence, simulation with support for various types of deformable objects is a much needed aid that can help to speed up experimentation with novel types of tasks, perception and manipulation algorithms. Interest in this direction has been indicated several recent workshops for modeling and manipulation of deformable objects, held at the leading robotics conferences~\cite{rss2021workshop, icra2021workshop, iros2020workshop}. A major obstacle is that the more advanced simulators, which support a broader range of objects and types of deformation, are difficult to tune manually. Hence, the community indicated the need for automated ways to find parameter estimates that make simulation behave stably, and make the behavior of the deformables resemble that of the real-world objects.

A number of existing methods for simulation parameter inference consider the case of rigid objects~\cite{chebotar2019closing, ramos2019bayessim, barcelos2020disco, mehta2020active, muratore2021neural, hwasser2020variational} and assume access to low-dimensional state, such as object poses.
One could argue that methods developed for rigid objects can be applicable to the case of deformable objects. For example, the recently popular keypoint extraction methods can generalize to the case of objects that are somewhat deformable, but still mostly maintain their shape (e.g. plush toys~\cite{florence2018dense}, flexible shoes~\cite{manuelli2019kpam}). However, most of these algorithms would not be applicable to the case of highly deformable fabrics, ropes and cables, where the object does not ever return to a canonical shape during manipulation. Such cases could benefit from new techniques that emerge in the machine learning community, e.g.~\cite{kulkarni2019unsupervised, li2020causal}. However, in this work we show that existing parameter inference methods need to be extended to work effectively with such learned state representations of deformables.

\subsection{Probabilistic Parameter Inference}

A recent survey~\cite{arriola2020modeling} includes an overview of methods for parameter estimation for simulators and models of deformable objects. These include direct error minimization and parameter estimation techniques that assume access to a direct way to compare the desirable and achieved deformation. To be applicable to a real-to-sim problem, this would require careful measurement of the deformation of the real deformable object, which can be intractable in many real-world scenarios. The alternatives that relax this assumption include exhaustive/random search and genetic algorithms, which require a large amount of compute resources. The techniques based on neural networks could improve data efficiency, but most lack the ability to capture uncertainty and only aid in producing a point estimate. This is problematic, because many currently available simulators for deformable objects are unable to produce behavior that exactly matches reality. Hence the need to combine such simulators with domain randomization techniques~\cite{matas2018sim, seita2019deep, andrychowicz2020learning}. This can ensure that control policies learned with the aid of simulation can handle a range of possible behaviors, instead of being narrowly focused on a mean estimate of the behavior.

BayesSim~\cite{ramos2019bayessim} is a likelihood-free method that has been applied to a variety of robotics problems~\cite{possas2020online, barcelos2020disco, matl2020stressd, matl2020inferring, mehta2020calibrating}. It offers a principled way of obtaining posteriors over simulation parameters, and does not place restrictions on simulator type of properties, i.e. can work with non-differentiable black-box models. BayesSim allows to infer multimodal posteriors with a {\em mixture density neural network\/} (MDNN)~\cite{bishop1994mixture, bishop2006pattern}, obtaining full covariance Gaussian components. Since Gaussian mixtures are universal approximators for densities~\cite{kostantinos2000gaussian, goodfellow2016deep}, given enough mixture components BayesSim posteriors can ensure sufficient representational capacity. Combining techniques based on neural networks and Bayesian inference allows BayesSim to be scalable and flexible in terms of its modeling capabilities. 

BayesSim performs probabilistic inference by considering a prior $p(\bth)$ over a vector of $D$ simulation parameters $\bth = [\theta_1, ..., \theta_D]$ and a derivative-free simulator used for obtaining trajectories of a dynamical system. Each trajectory ${\bx}^s$ is comprised of simulated observations for states $\bS=\{\bs\}_{t=1}^T$ of a dynamical system and the actions $\bA=\{\ba\}_{t=1}^T$ that were applied to the system. BayesSim then collects a few observations from the real world, e.g. a single trajectory ${\bx}^r$ and uses it to compute the posterior $p\Big(\bth \Big| \big\{{\bx}_{(i)}^s\big\}_{i=1}^N, {\bx}^r\Big)$. Instead of assuming a particular form for the likelihood and estimating $p(\bx | \bth)$, BayesSim approximates the posterior by learning a conditional density $q_{\phi}(\bth | \bx)$, represented by an MDNN with weights $\phi$. The posterior is then:
\begin{align}
\label{eq:bsiminfer}
\hat{p}(\bth | \bx \!=\! {\bx}^r) \propto p(\bth) / \tilde{p}(\bth) q_{\phi}(\bth | \bx \!=\! {\bx}^r), 
\end{align}
with an option for a proposal prior $\tilde{p}(\bth)$ used to collect simulated observations to train the conditional density. 

In previous works, BayesSim first summarized simulated and real trajectories by extracting the sufficient statistics:
\begin{align}
\varphi(\bS,\bA) = \big( \{\langle \btau_i, \pmb{a}_j \rangle\}_{i,j=1}^{D_s,D_a}, \mathbb{E}[\btau], \mathbb{V}ar[\btau] \big),
\end{align}
where $\btau = \{\bs_t - {\bs}_{t-1}\}_{t=1}^T$ contains the trajectory state differences, $D_s,D_a$ are the state and action dimensionalities, $\langle \cdot , \cdot \rangle$ denotes a dot product between the state and action feature vectors. Works that applied BayesSim to low-dimensional states (e.g. robot joint angles, object poses and velocities) used these states directly. \cite{matl2020inferring} applied BayesSim to scenarios with granular media and developed domain-specific summary statistics of depth images of granular formations, such as dispersion and statistical dependence of the grain locations (mean, standard deviation, interquartile range, kurtosis, and distance correlation).
In this work, we present a novel methodology to perform inference using state trajectories directly from images, without the need to first extract sufficient statistics from trajectories.

\section{Our Approach : Bayesian Inference with Distributional RKHS Embeddings}
\label{sec:approach}

\subsection{BayesSim with a Distributional View of Deformables}

In this section we describe BayesSim for deformables. The method begins by first executing a trajectory with a robot manipulating a deformable object. For example, for the wiping scene, this could correspond to executing a simple horizontal motion, where the robot drags a cloth on a table surface. We record robot poses and RGB images of the scene. Our goal is to infer a  simulation parameter posterior, such that samples from it yield simulations that resemble reality in terms of deformable object motion.
We obtain simulated trajectories and RGB images from simulations of the scene with a deformable object. Simulation parameters are sampled from a uniform distribution on the 1st iteration. We use these initial real and simulated images to train keypoint extraction models. We extract the keypoints from simulated images and include them in the state observations:
\begin{align}
\bS = \{\bs\}_{t=1}^T: \bs_t = \big[ gripper\!\_pose, \!\  \bk_1,...,\bk_K \big],
\end{align}
where $gripper\!\_pose$ is the Cartesian pose of the gripper, and $\bk_1, ... , \bk_n$ are the extracted keypoints (2D in pixel coordinates or 3D in world frame if camera-to-robot transform is given). Following BayesSim, we obtain simulation training trajectories $\bx^{s}\!=\!\{\bs_t,\ba_t\}_{t=1}^T$, where $\ba_t$ are the robot actions, e.g. target Cartesian gripper poses, joint angles or velocities (depending on the desired control mode). We use a set of simulated trajectories to learn a conditional density $q_{\phi}(\bth | \bx)$ represented by an MDNN with weights $\phi$. We then obtain the posterior $\hat{p}_1(\bth|\bx=\bx^r)$ using the real trajectory $\bx^r$ and Equation~\ref{eq:bsiminfer}. The above constitutes one iteration. We obtain a new set of simulation trajectories by sampling simulation parameters from the approximate posterior $\hat{p}_1$, then repeat the above steps to obtain $\hat{p}_2$ as posterior for the 2nd iteration, and so on.

To guarantee that our state representation has favorable properties, both in terms of theory and practice, we propose to transform the part of the state that contains keypoints $\bk_1,...,\bk_K$. Our insight is that the keypoints can be viewed as noisy samples from a probability distribution that has support on the surface of the deformable object in the scene. The full state of the deformable is unobservable due to occlusions by other objects as well as self-occlusions. Moreover, keypoint extraction methods do not guarantee ordering, and do not guarantee placement of the keypoints on consistent parts of the object. 
Nonetheless, our insight of treating them as samples from the distribution that captures the state of the deformable allows us to overcome these shortcomings. Furthermore, this distributional treatment yields a method that is robust to noise by construction, and can benefit from principled theoretical tools for analysis and interpretation.

\subsection{An Intuitive Explanation of Kernel Mean Embeddings}

Kernel mean embeddings allow to map distributions into infinite dimensional feature spaces~\cite{song2013kernel}. 
Conceptually, they are able to represent probability distributions without loss of information in a feature space that is more amenable to mathematical operations. A desirable property that a kernel mean embedding satisfies is that it can recover expectations of all the functions in a given reproducing kernel Hilbert space $\mathcal{F}$. This property of the mean embedding map $\mu_X \in \mathcal{F}$ is technically stated as:
\begin{align}
\mathbb{E}_X[f(X)] = \langle \mu_X , f \rangle, \ \forall f \in \mathcal{F}.
\end{align}
To get an intuition for why this is useful, consider the following example: if an RKHS $\mathcal{F}$ is sufficiently large, e.g. includes all monomials $\{X, X^2, X^3, .... \}$, then we can obtain all the moments of the distribution of $X$ by simply taking dot products with $f \in \mathcal{F}$.
In the above, $X$ is a random variable with domain $\Omega$ and distribution $P(X)$, $\mathcal{F}$ is an RKHS on $\Omega$ with a kernel $k (x, x')$. 
This means that $\mathcal{F}$ is a Hilbert space of functions \mbox{$f: \Omega \rightarrow \mathbb{R}$} with the inner product $\langle \cdot, \cdot \rangle_{\mathcal{F}}$. 

Kernel slices $k (x, \cdot)$ are functions in $\mathcal{F}$ that satisfy the \textit{reproducing property}: $\big\langle \!\ f(\cdot) \ , \ k(x, \cdot) \!\  \big\rangle_{\mathcal{F}} = f(x)$. A kernel slice $k (x, \cdot)$ can be viewed as an implicit feature map $\phi(x)$ with $k(x, x') = \big\langle \ \phi(x) \ , \ \phi(x') \ \big\rangle_{\mathcal{F}}
$.
\cite{fukumizu2007kernel} shows that commonly used kernels are indeed sufficiently large. These \textit{characteristic kernels} ensure that $\mu_X$ mapping is injective, hence embedding $P(X)$ as $\mu_X$ does not lose information about $P(X)$. The RBF kernel $k(x, x') = \exp(-\sigma ||x-x'||_2^2)$ is the most widely used kernel. The fact that it can be formed by taking an infinite sum over polynomial kernels connects to our `moments of distributions' example above.

The Riesz representation theorem states that $\mu_X$ in Equation~4 exists and is unique. Using the reproducing property and linearity of integration, an explicit formula for $\mu_X$ can be obtained:
$\mu_X = \mathbb{E}[\phi(X)] = \int_{\Omega} \phi(x) \ dP(x)$ (after simplifying).
This suggests defining the empirical kernel embedding using i.i.d. samples $x^{(1)}, ..., x^{(N)}$ from $P(X)$ as:
\begin{align}
\hat{\mu}_X := \tfrac{1}{N} \sum_{n=1}^N \phi\big(x^{(n)}\big).
\end{align}
\cite{smola2007hilbert} justifies the above choice by showing that $\hat{\mu}_X$ converges to $\mu_X$ as $O(1/\sqrt{N})$, independent of the dimensionality of $X$, which avoids the curse of dimensionality. Instead of dealing with infinite-dimensional implicit maps $\phi(x)$, applying the \textit{kernel trick} allows to operate with the finite-dimensional Gram matrix $K: K_{ij} = k\big(x^{(i)}, x^{(j)}\big)$, $i,j=1...N$.

\subsection{RKHS-Net Layer for Distributional Embeddings}
\label{sec:rkhsnet}

\begin{figure}[t]
\centering
\vspace{6px}
\includegraphics[width=1.0\linewidth]{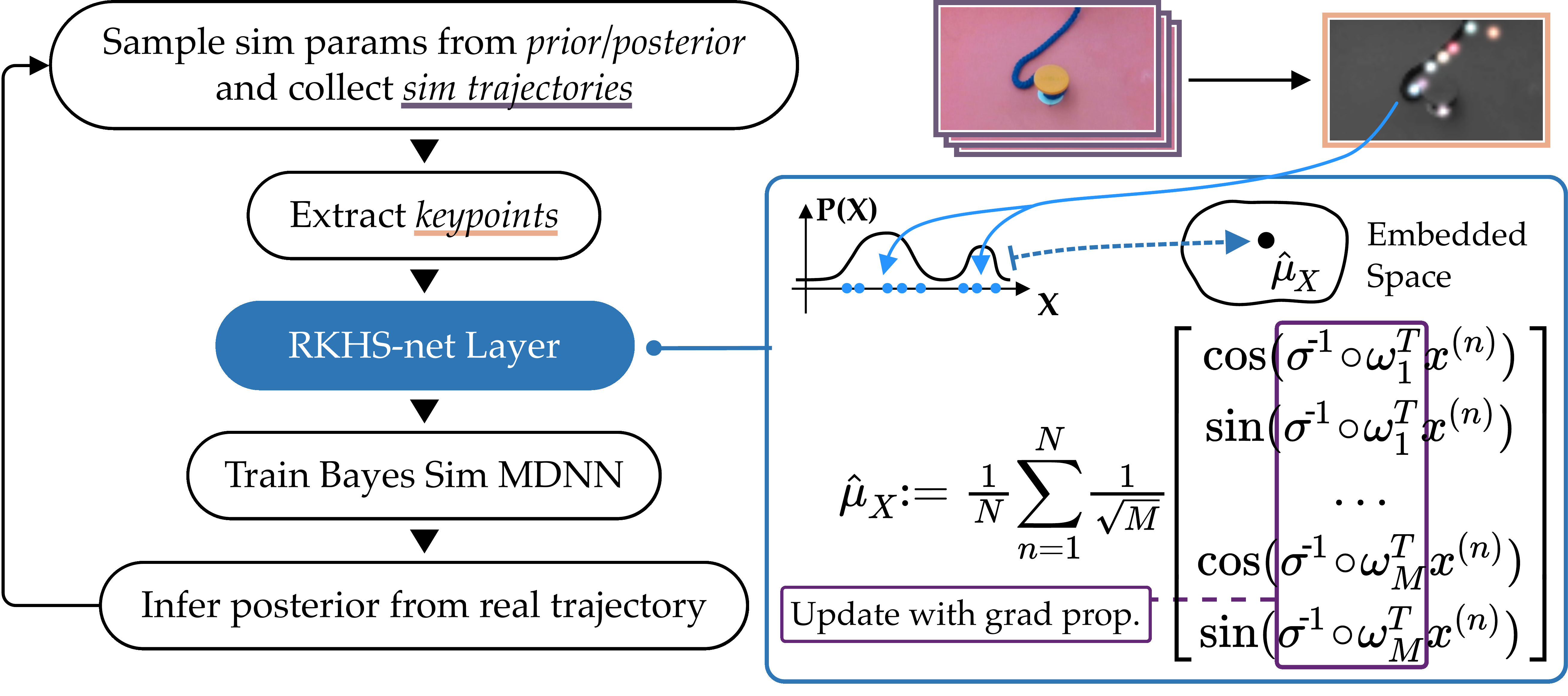}
\vspace{-17px}
\caption{An overview of our \textit{BayesSim-RKHS} method, with a focus on the proposed RKHS-net layer, shown within the blue rectangle. The RKHS-net layer can take samples from any distribution as inputs, and in this work we compute the distributional embedding for the keypoints $\bk_1,...,\bk_K$.}
\label{fig:meanembed}
\vspace{-10px}
\end{figure}

For scalability reasons, we can avoid the computation of the Gram matrix by approximating the kernel function by its inner product:
\begin{align}
k(x,x') = \big \langle \ \phi(x) \ , \  \phi(x') \ \big \rangle_{\mathcal{F}} \! \approx \!\ \hat{\phi}(x)^T \hat{\phi}(x'), 
\end{align}
where $\hat{\phi}(x)$ is a finite dimensional approximation of $\phi(x)$, known as \textit{random Fourier features}~\cite{rahimi2007random, rahimi2008weighted}.
Following the derivation in~\cite{rahimi2007random}, we first employ \textit{Bochner's theorem}, which states that any continuous shift-invariant kernel  $k(x,x') := k(x-x')$ can be represented in terms of its Fourier transform:
\begin{align*}
k(x-x') = \int p(\omega) \!\ \exp\big(i \omega^T (x-x')\big) d\omega,
\end{align*}
where $p(\omega)$ is the spectral density corresponding to kernel $k(x-x')$.  
For real a real-valued kernel $k(\cdot, \cdot)$, the right-hand side can be written without the imaginary part as $\mathbb{E}_\omega \big[\cos\big(\omega^T (x-x')\big)\big]$. This expectation can be approximated with a Monte Carlo estimate yielding:
\begin{align*}
&k(x-x') \approx \frac{1}{M} \sum_{m=1}^{M} \cos \big(\omega_{m}^T x - \omega_{m}^T x'\big) 
=
\hat{\phi}(x)^T \hat{\phi}(x'),
\\
&\hat{\phi}(x)^T \!\!=\!\!
\tfrac{1}{\sqrt{M}} 
\Big[ \cos(\omega_1^T x), \sin(\omega_1^T x), ..., \cos(\omega_M^T x), \sin(\omega_M^T x) \Big]
\end{align*}
When $\omega \!\sim\! \mathcal{N}(\pmb{0}, I)$, the above approximates an RBF kernel with $\sigma\!\!=\!\!1$. More generally, when $\omega \!\sim\! \mathcal{N}(\pmb{0}, \sigma I)$ this approximates an RBF kernel with a hyper-parameter $\sigma$. The \textit{frequencies} $\omega$ are usually sampled randomly, yielding components of the form $\cos(\sigma^{-1} \!\circ\! \omega_M^T x)$, $\sin(\sigma^{-1} \!\circ\! \omega_M^T x)$.
\cite{rahimi2007random} provides approximation bounds and further analysis of the random Fourier features (RFF) approximation. 

We propose to use the RFF feature approximation to construct the mean embedding for the part of the state that benefits from the distributional representation. In the current work this includes the keypoints $\bk_1,...,\bk_K$. Though, in general, the proposed approach is not limited to handling keypoint representations, and can embed any distributional part of the state. Furthermore, we propose to integrate this into the overall learning architecture in a fully differentiable manner. We accomplish this by constructing a
neural network layer that obtains random samples for $\omega$, and then propagates the gradients through to adjust them during training. We also propagate the gradients through $\sigma$. Figure~\ref{fig:meanembed} illustrates this.

\subsection{Keypoint Extraction Modules}
\label{sec:keypointtrain} 

\begin{figure}[b!]
\vspace{-10px}
\centering
\includegraphics[width=1.0\linewidth]{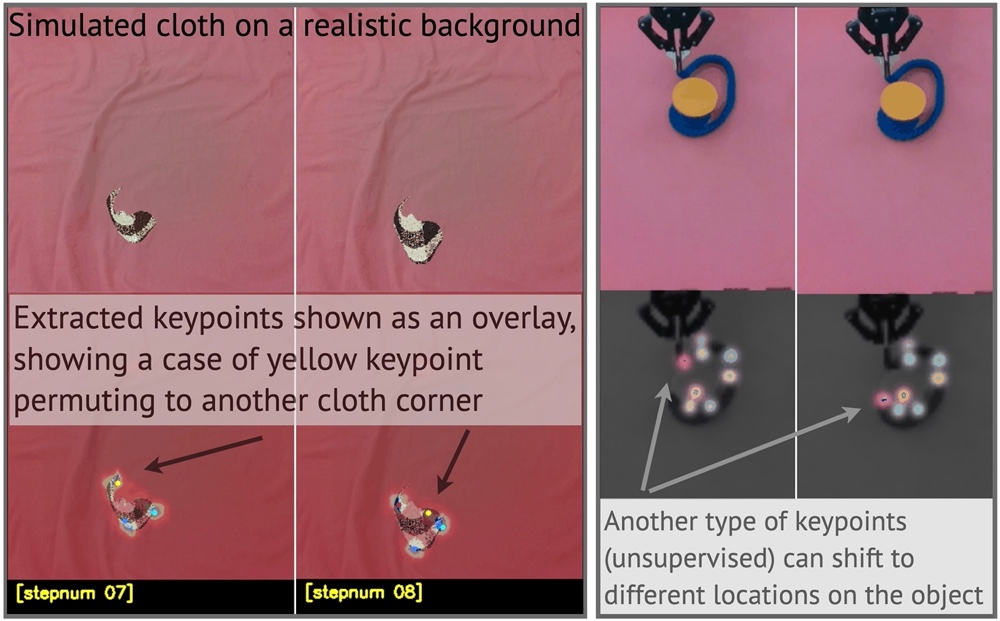}
\vspace{-15px}
\caption{Left: keypoints from the supervised approach~\cite{sundaresan2020untangling} appear close to the desired corner regions, despite deformations. They tend to track consistent locations, but the method does not aim to guarantee this. Right: example results from our adaptation of the unsupervised method from~\cite{kulkarni2019unsupervised}.}
\label{fig:keypoints}
\end{figure}

We now describe two keypoint extraction methods: one is a data-efficient supervised method, for which the user needs to annotate a small set of images to indicate the desired locations for the keypoints~\cite{sundaresan2020untangling}; the other is unsupervised, and only needs unlabeled RGB frames for training~\cite{kulkarni2019unsupervised}.

\cite{sundaresan2020untangling} is a recent method designed for learning features to help refine coarse keypoint prediction to a precise consistent location on an object. This method has been shown to work well as part of a larger framework aimed to solve the perception, planning and manipulation challenges for the task of untangling knots.
The aim is to learn semantic keypoints that roughly capture the state of the deformable objects. For scenarios with cloth we annotate the corners of the cloth to indicate the desired areas where the algorithm should learn to place the keypoints. For ropes it is less obvious what the `best' location for placing a keypoint should be. Hence, we make a simple choice of spacing the keypoints uniformly along the rope. 
Using these annotated images as RGB image observations, we learn a mapping $f: \R^{W \!\!\times\! H \!\times\! 3} \rar R^{W \!\!\times\! H \!\times\! 4}$, where each channel of the output represents a 2D heatmap for one keypoint. Given 250 images (125 simulated, 125 real), we annotate four task-relevant keypoints on each image. Then, we apply affine, lighting, and color transformations to augment the dataset, obtaining the overall augmented dataset size of 3000 images. A network with a ResNet-34 backbone is then trained to predict 2D Gaussian heatmaps centered at each keypoint. 
After training, the positions of the keypoints are predicted as argmax over each channel heatmap.

We aim for our overall approach to effectively handle noisy outputs that unsupervised approaches can yield as well. For this, we adapt an unsupervised keypoint extraction approach based on the Transporter architecture~\cite{kulkarni2019unsupervised}. This method takes as input RGB images $x_{src}$ and $x_{tgt}$. A convolutional neural network (CNN) and a keypoint detection network encode the input to the spatial feature map $\Phi(x)$ and a keypoint network $\Psi(x)$. Then, a `transport' operation is performed to modify the feature map of the input image,  such that source features at the location of the source image keypoints are subtracted out, while target features at the location of the target image keypoints are pasted in: $\hat{\Phi}(x_{src}, x_{tgt}) = (1 - \mathcal{H}_{\Psi(x_{src})}) \cdot (1 - \mathcal{H}_{\Psi(x_{tgt})}) \cdot \Phi(x_{src}) + \mathcal{H}_{\Psi(x_{tgt})} \cdot \Phi(x_{tgt})$.
Here, $\mathcal{H}$ denotes the mapping from keypoint coordinates to Gaussian heatmaps. Then, a decoding CNN reconstructs the target image from the transported feature map. The training is guided only by the reconstruction loss.
We extend this approach to ensure that the keypoint network $\Psi(x)$ allocates keypoints to the manipulated objects, as opposed to placing them on the robot. The pose of the robot and its geometry (mesh) are usually known to high precision. Hence, there is no benefit in tracking the motion of the robot from the RGB images. 
We obtain the robot mask (the part of the image that has robot in the foreground) by using depth filtering methods on the depth readings that we acquire from an RGBD camera mounted in the workspace.
We mask out the areas with the robot from the reconstruction loss during training, and this helps the method focus on the regions with the deformable object. The right side of Figure~\ref{fig:keypoints} shows an example of the keypoints we obtain with this approach.

\section{Experiments}
\label{sec:experiments}

In the following sub-sections, we first describe the scenarios we consider, then explain our hardware setup and evaluation strategy, then illustrate the results of real-to-sim experiments using two types of keypoint extraction methods.

\begin{figure}[t]
\vspace{5px}
\centering
\includegraphics[width=1.0\linewidth]{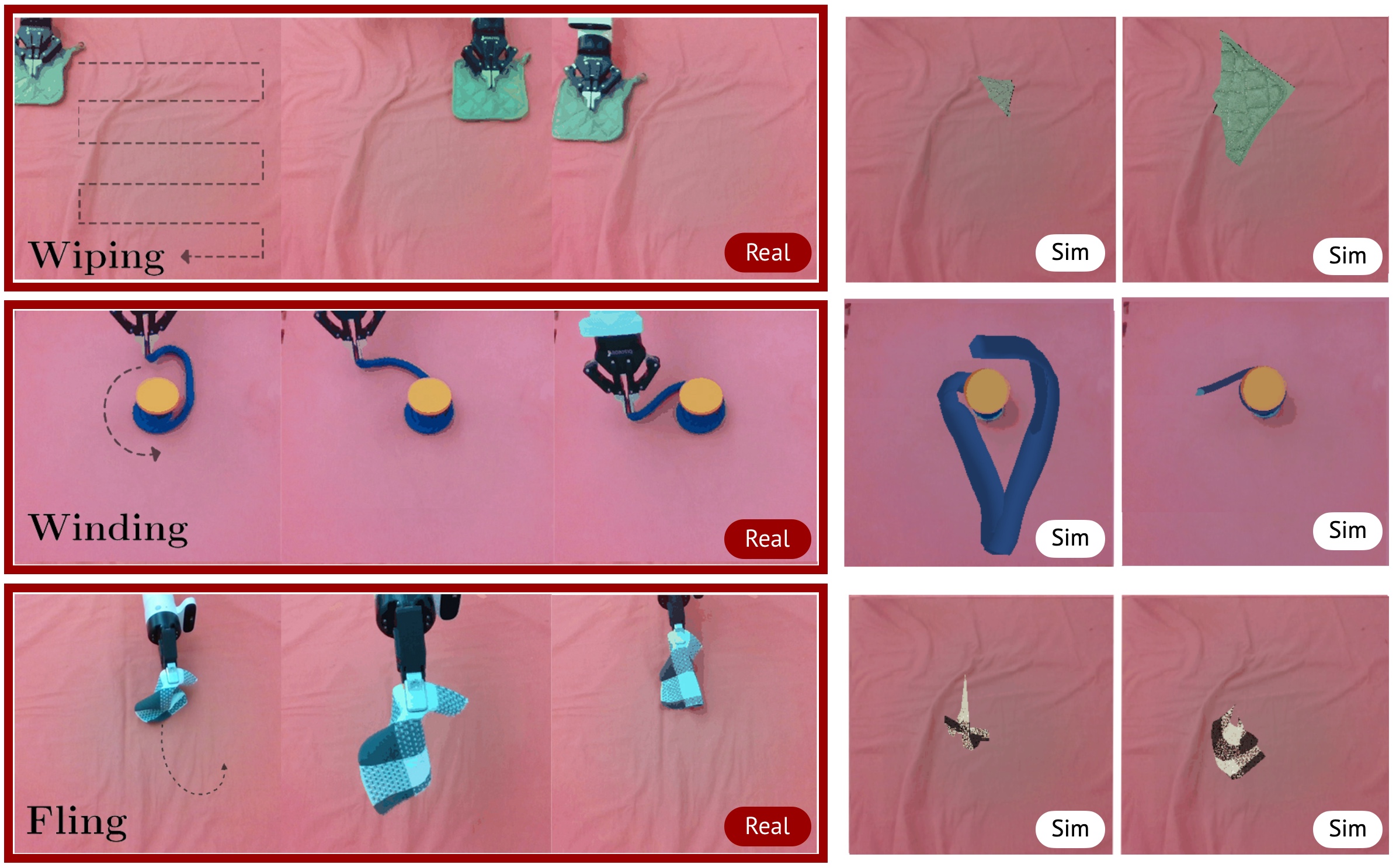}
\vspace{-20px}
\caption{Left: scenarios we consider in our hardware experiments. Right: examples of simulation with various parameters; the initial parameter ranges are wide, which yields both realistic and unrealistic behavior, as expected.}
\label{fig:scenarios}
\vspace{-10px}
\end{figure}

\subsection{Description of Real and Simulated Scenarios.}

For our experiments we consider three scenarios that involve deformable objects with various levels of deformation.
The first is a wiping scenario, where a robot manipulates a thick cloth to wipe a table surface. 
This scenario aims to test the case when capturing the state of the real object is tractable, but finding an appropriate simulation posterior is challenging.
The real wiping cloth is clearly visible in most frames, and undergoes only small deformations. In contrast, simulated cloth can be highly flexible, and easily crumbles when medium-to-low bending and elastic stiffness simulation parameters for the cloth are sampled together with medium-to-high values for the friction parameter. These cases are frequent in the initial uniform simulation parameter samples.

In the second scenario the robot winds a highly flexible rope around a spool. This scenario presents a challenge for the keypoint extraction methods, since there are no obvious canonical locations for keypoints. Furthermore, parts of the rope are occluded by the spool.

In the third scenario the robot flings the cloth up, then lowers it down and drags it on the table surface. 
This scenario is challenging for perception, both for real images and simulation, since the cloth is highly flexible and not fully visible at any point. With medium-to-high friction, the ends of the cloth spread out on the table, but the top corners of the cloth remain obscured due to self-occlusion.

In all of these scenarios we infer a joint posterior for bending stiffness, elastic stiffness, friction, and the scale/size of the deformable object. The rest of the simulation parameters are left as defaults, and are the same across all the scenarios. We use the PyBullet simulator~\cite{coumans2019}, with the Finite Element Method (FEM) option for simulating deformable objects. While FEM can be computationally expensive and precise in general, the default settings we use in PyBullet prioritize speed over fidelity to obtain faster-than-realtime simulation. Figure~\ref{fig:scenarios} shows visual examples of our scenarios. 

In this work, we focus on inference of posteriors of physical simulation parameters, and do not explore the aspect of a large mismatch in camera perspective or visual appearance of the scene. Hence, we make simulation environments that approximately match the visual appearance, which is easy to achieve in the PyBullet simulator. We load a realistic background and texture for the deformable object, and approximately match the camera pose. Addressing a large visual gap for a simulator that cannot import custom visual elements can be done by either training keypoint methods with heavy visual domain randomization, or by exploring novel differentiable rendering techniques. We leave these directions for future work. In this work we also do not focus on the aspect of simulating grasping of the deformable object with high fidelity. In our simulated environments we attach a simple grasp anchor to the cloth/rope objects, instead of simulating the interaction between the gripper's fingers and the thin-shell deformable object. SoftGym~\cite{lin2020softgym} is a recent example of a suite of simulation environments geared towards sim-to-real that adopts a similar approach for a subset of environments.

\begin{figure}[t]
\vspace{10px}
\centering
\includegraphics[width=1.0\linewidth]{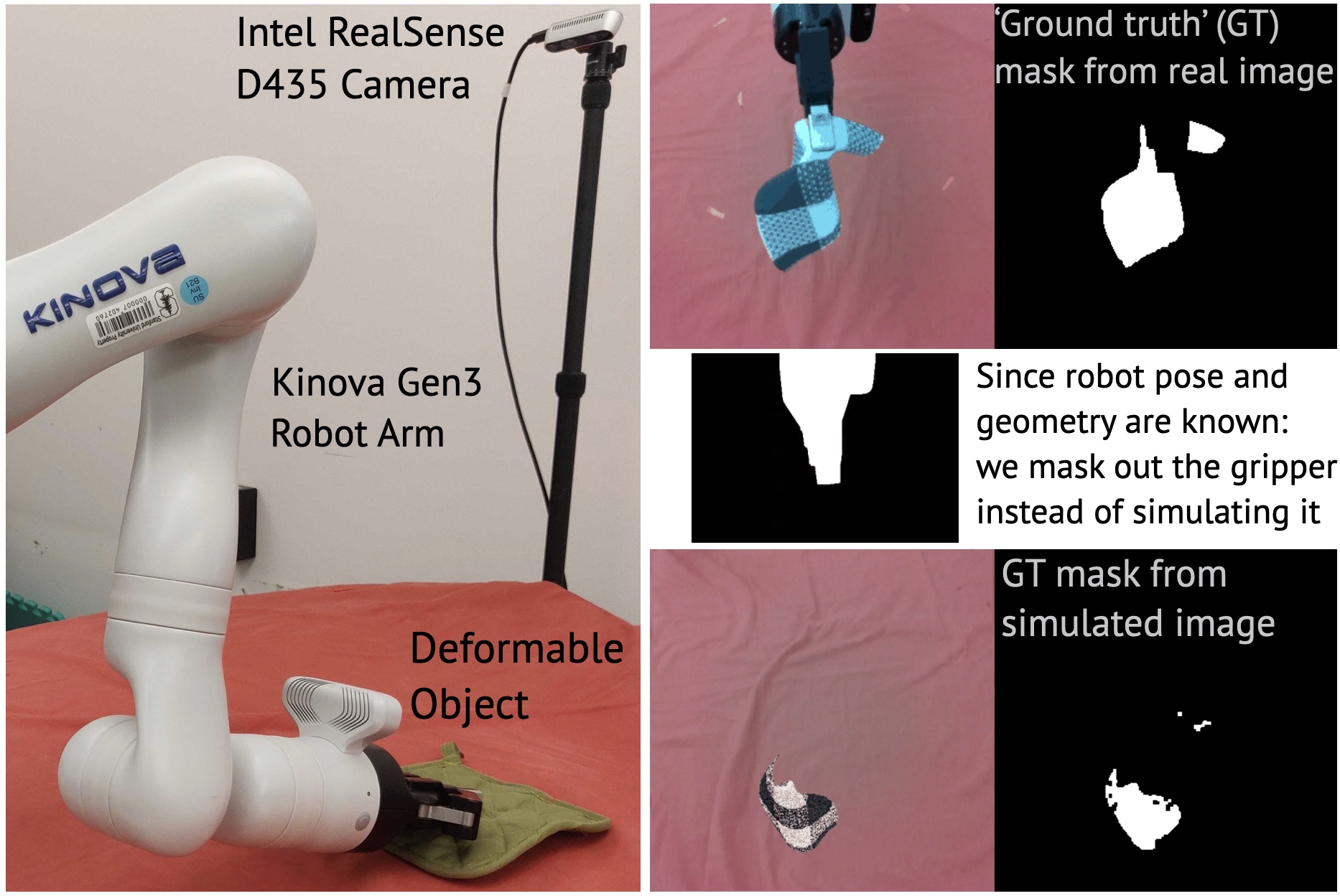}
\vspace{-20px}
\caption{Left: our hardware setup. Right: visualization of our evaluation strategy to measure the alignment between the behavior of the real and a simulated deformable objects. The `ground truth' masks are only used for evaluation, and are not given to any of the algorithms we compare.}
\label{fig:hw_setup}
\vspace{-10px}
\end{figure}

\subsection{Hardware Setup and Evaluation Methodology}

Our hardware setup includes a Kinova Gen3 7DoF robot arm with a Robotiq 2F-85 gripper, and an Intel RealSense D435 camera. The camera provides RGB image data during experiments and is positioned to view the table surface. As image resolution, we use $320 \times 320$ pixels in all our experiments. To execute the desired robot trajectories we use velocity control in the Cartesian (end-effector) space, using the high-level control interface provided by Kinova. The left side of Figure~\ref{fig:hw_setup} illustrates our workspace.

To evaluate the performance of our approach and baselines, we measure alignment between the motion of the real and simulated deformable object. To localize the deformable objects in the scene we construct masks based on color filters defined in HSV space. The mask extracted for the real object constitutes the `ground truth' region. Note that this ‘ground truth’ (GT) is only used for evaluation purposes and is not given to any of the algorithms we compare which only use keypoints as input. We construct a mask for the simulated object as well, then compare the two masks. To compare the two masks, we use a bidirectional Chamfer distance, which is a common metric for measuring differences between unordered point clouds. To obtain trajectories for evaluation, we command the same trajectory on the real robot and in simulation, obtain the frames from the real and simulated cameras, then compute the Chamfer distance. The mean of this distance across all timesteps constitutes our evaluation metric that quantifies alignment between the real and simulated deformable object. On the $y$ axis in our plots, we refer to this metric  as the ``distance to `ground truth' state''.

\subsection{Hardware Results for Real-to-sim}

\begin{figure}[t]
\vspace{10pt}
\centering
\includegraphics[width=0.9\linewidth]{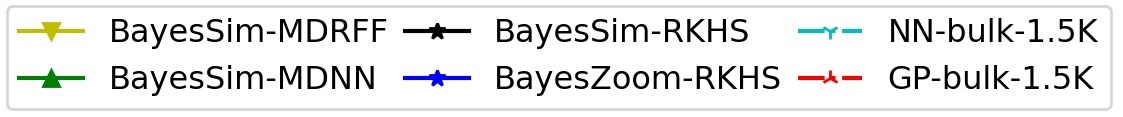}
\includegraphics[width=0.495\linewidth]{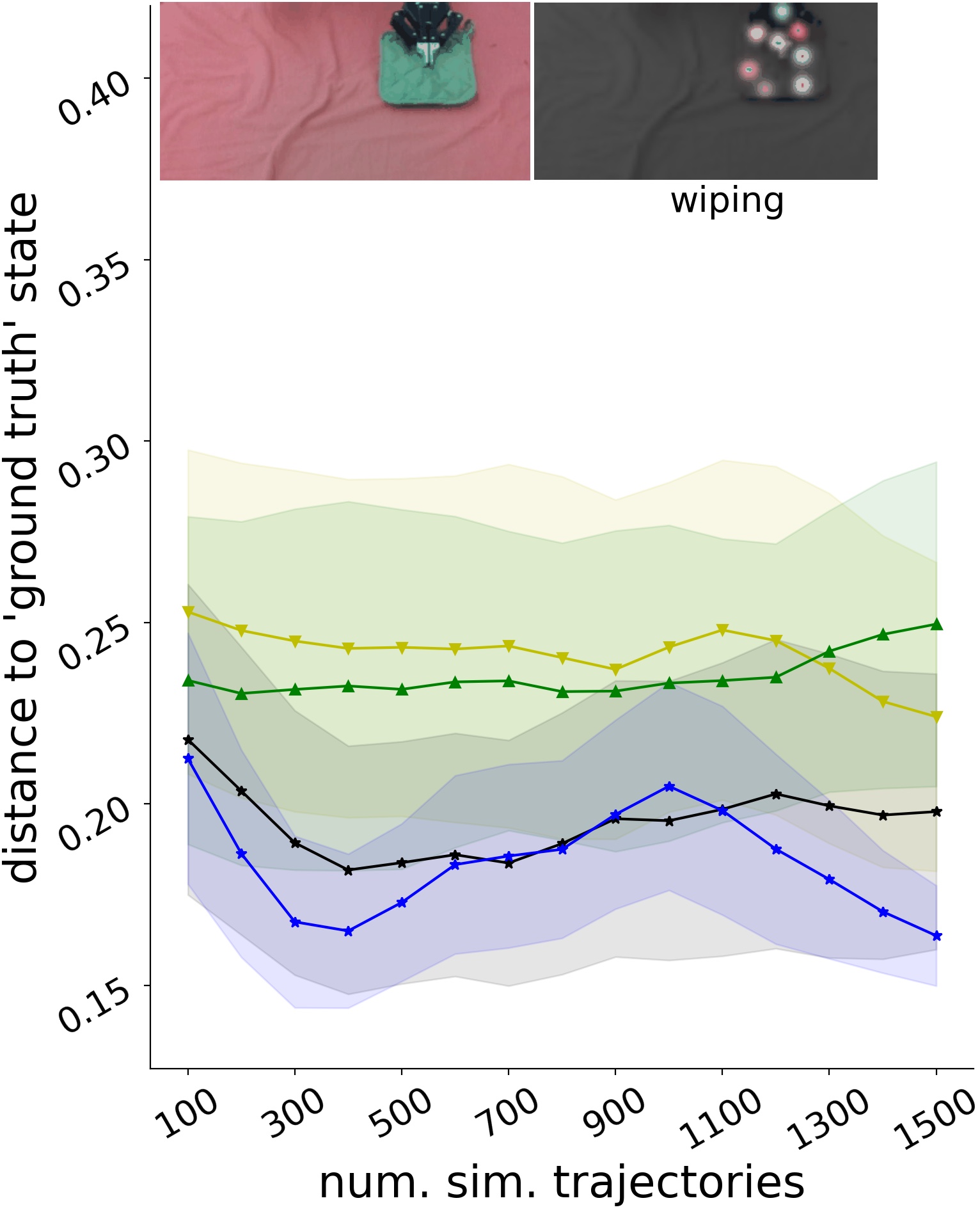}
\includegraphics[width=0.475\linewidth]{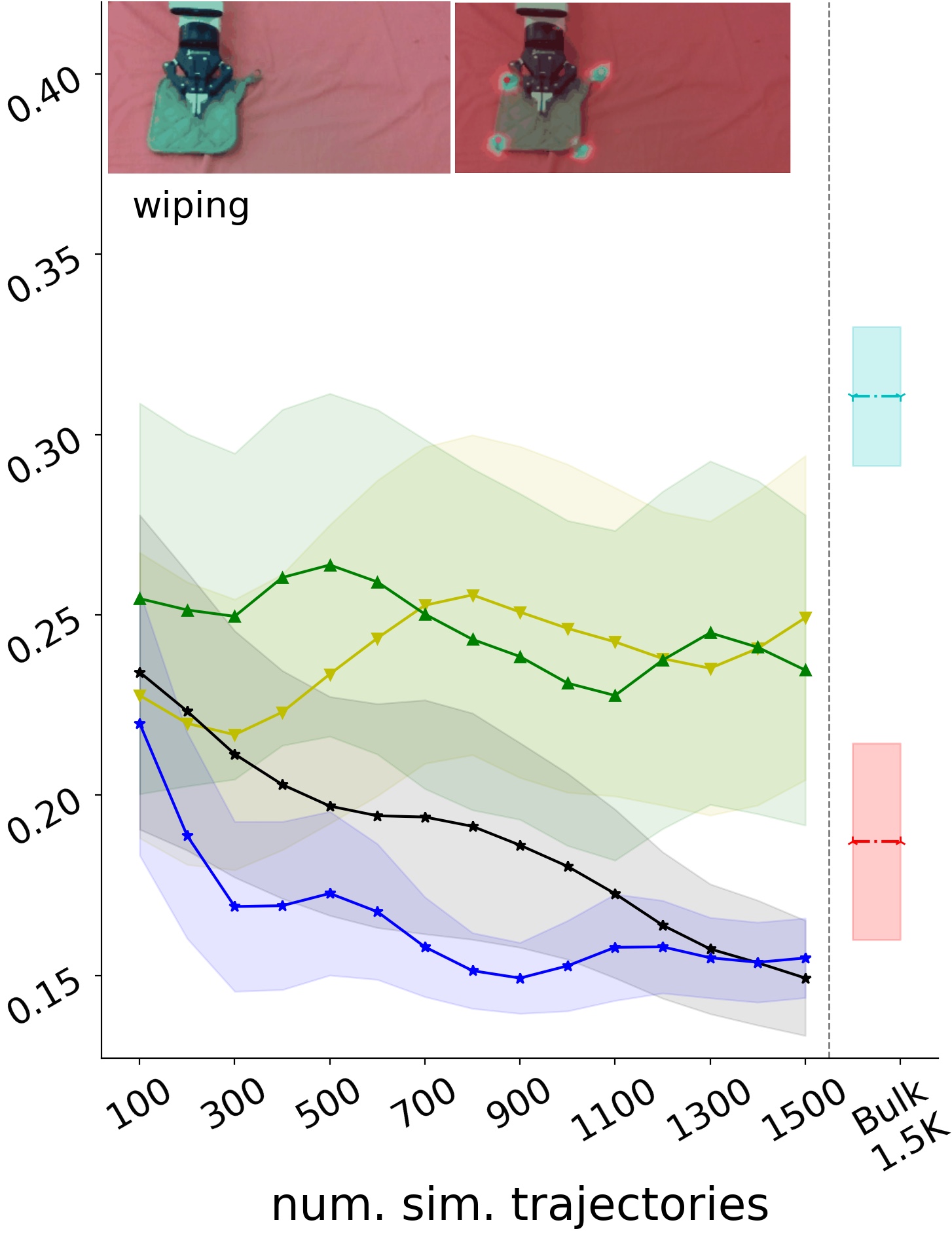}
\vspace{-11px}
\caption{Hardware results for the wiping task. Left: using unsupervised keypoints. Right: using supervised keypoints.}
\label{fig:hw_results_wiping}
\vspace{-12px}
\end{figure}

For evaluation we compare the following 6 methods:

\textit{BayesSim-MDNN} is the original BayesSim method in~\cite{ramos2019bayessim}. 

\textit{BayesSim-MDRFF} extracts RFF features from trajectories before passing the data to the mixture density network for training. This variant has been used in experiments with inferring material properties of granular media~\cite{matl2020inferring} and showed strong performance on that challenging task. One key aspect to note is that~\cite{matl2020inferring} domain-specific features, as we described in the background in Section~\ref{sec:background}. We aim to avoid designing domain-specific features for manipulation with deformables, since it is not plausible to presuppose that a single feature extraction technique could perform well across various scenarios/tasks and various kinds of objects and deformation types.

\textit{BayesSim-RKHS} is the method we propose in Section~\ref{sec:approach}.

\textit{BayesZoom-RKHS} is a variation of our method, which we test to verify that our distributional embedding of keypoints can still work with a simpler variant of BayesSim. This variant uses a \mbox{3-layer} fully-connected neural network instead of the mixture density network. The network learns to predict the mean of a unimodal Gaussian posterior. We use the L1 error on the validation set as a way to compute an estimate of the square root of the variance. We refer to this method as \textit{BayesZoom}, since it retains the core idea of shifting the posterior to re-sample simulated trajectories from the more promising regions of the search space (i.e. it `zooms' in on the useful parts of the space). \textit{BayesZoom-RKHS} incorporates our idea of using distributional embedding for keypoints.

All of the above algorithms are sequential: they shift the posterior after each iteration, then sample simulation data from the new posterior on the subsequent iteration. We collect 100 simulation trajectories on each iteration, and retain all the data from the previous iterations for training. Hence, after 15 iterations these algorithms would collect a total of 1.5K simulation trajectories overall. To compare this sequential vs batch/bulk training, we test the following two `bulk' algorithms that collect 1.5K trajectories sampled from the uniform prior, then train on this dataset.

\textit{NN-bulk-1.5k} is a batch method that trains a fully connected neural network on 1.5K simulated trajectories.

\textit{GP-bulk-1.5k} performs batch Gaussian process regression.

In all NN-based algorithms, we use 3-layer fully connected neural networks as the `backbone' with 1024 units in each layer. We use Adam optimizer with a learning rate of \mbox{$1e\text{-}6$}. In our experience, the small learning rate helps NN-based approaches to learn from noisy data. For Gaussian process regression we use GPyTorch+BOTorch with automatic hyperparameter optimization. 
For keypoint extraction, we experimented with using 8 and 4 keypoints with the unsupervised method. For the supervised approach we used 4 keypoints to minimize the time spent on labeling the data.

\begin{figure}[t]
\vspace{10pt}
\centering
\includegraphics[width=0.9\linewidth]{img/legend.jpg}
\includegraphics[width=0.505\linewidth]{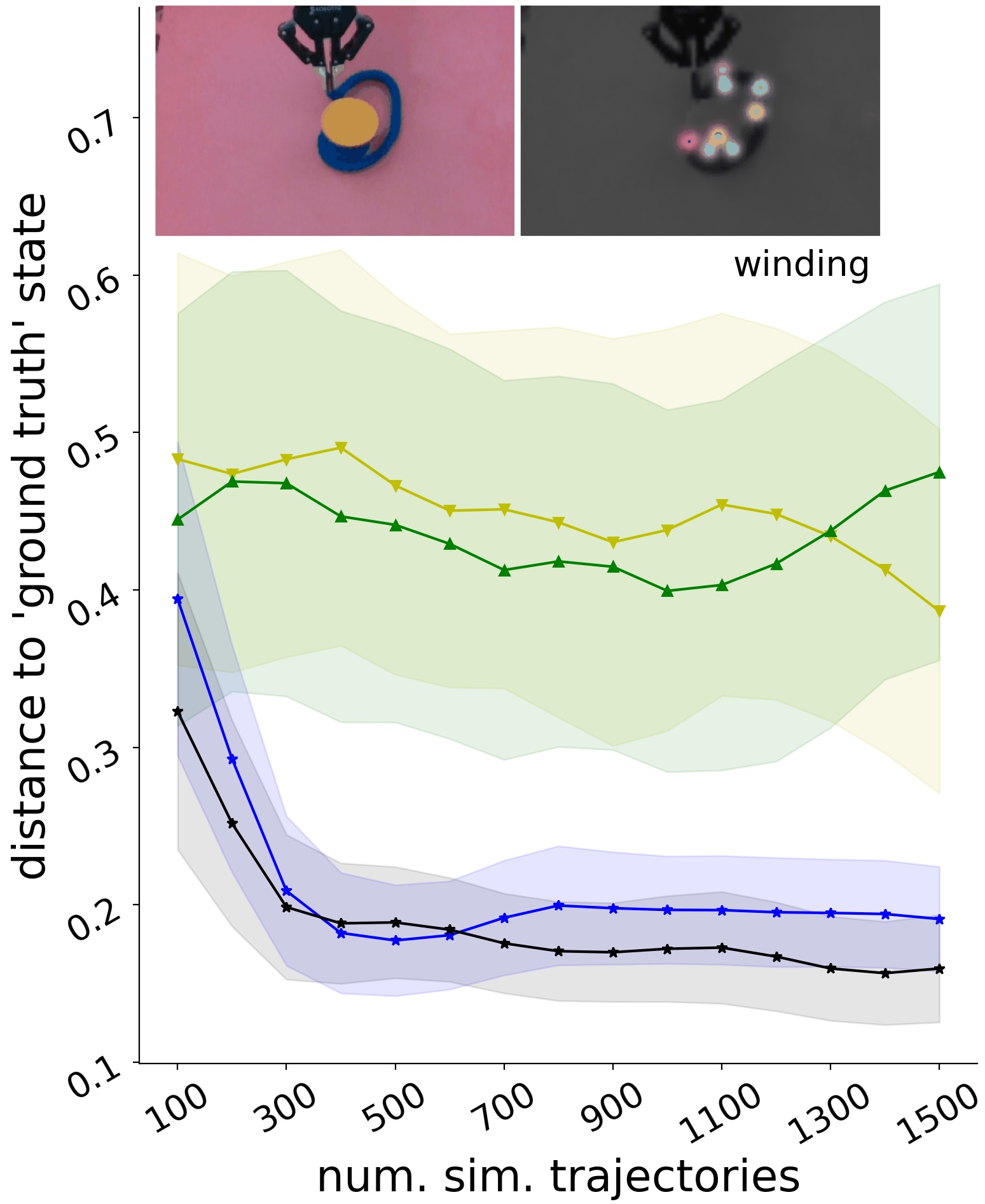}
\includegraphics[width=0.480\linewidth]{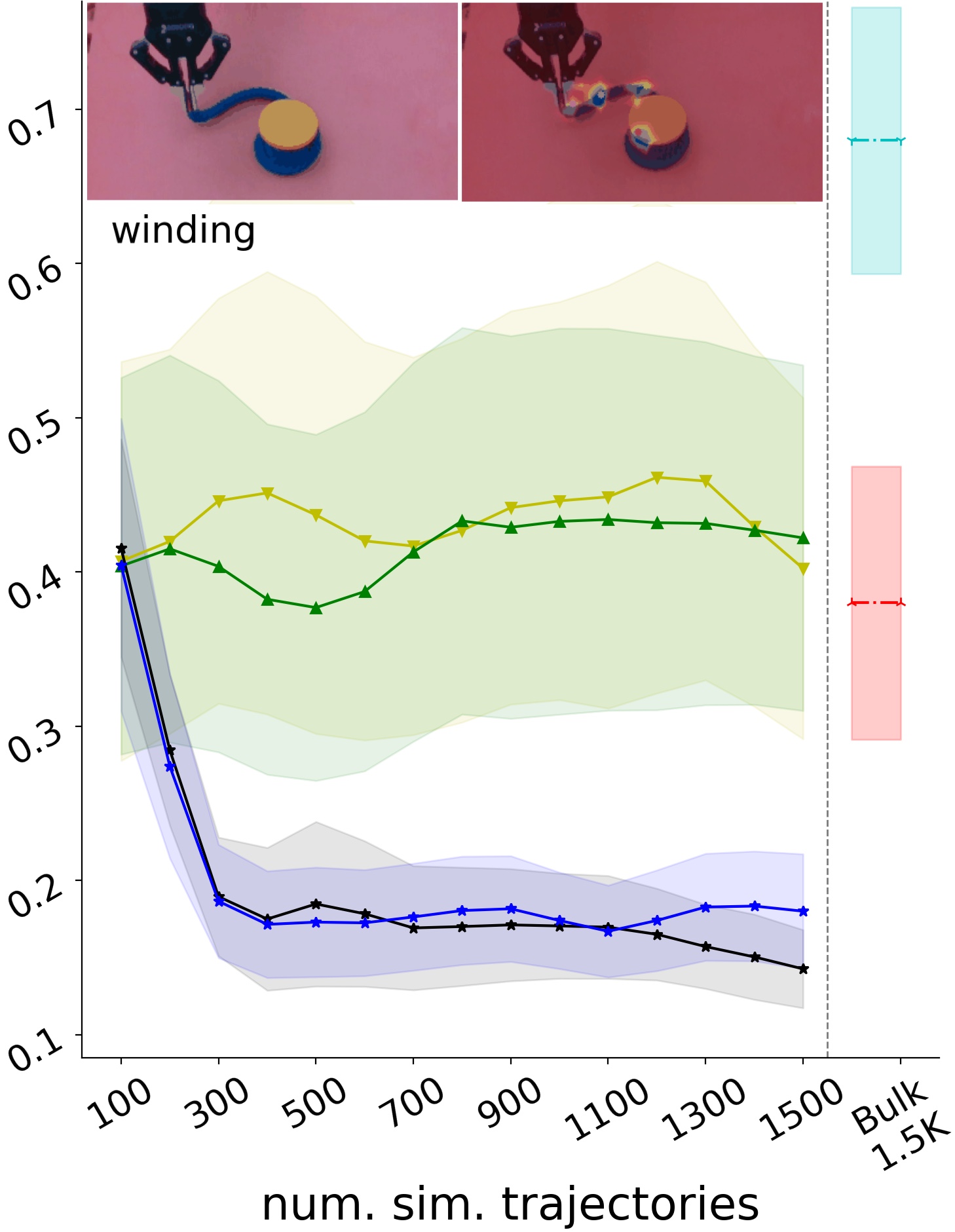}
\vspace{-18px}
\caption{Hardware results for the winding task. Left: using unsupervised keypoints. Right: using supervised keypoints.}
\label{fig:hw_results_winding}
\vspace{-10px}
\end{figure}

\begin{figure}[b]
\centering
\includegraphics[width=1.0\linewidth]{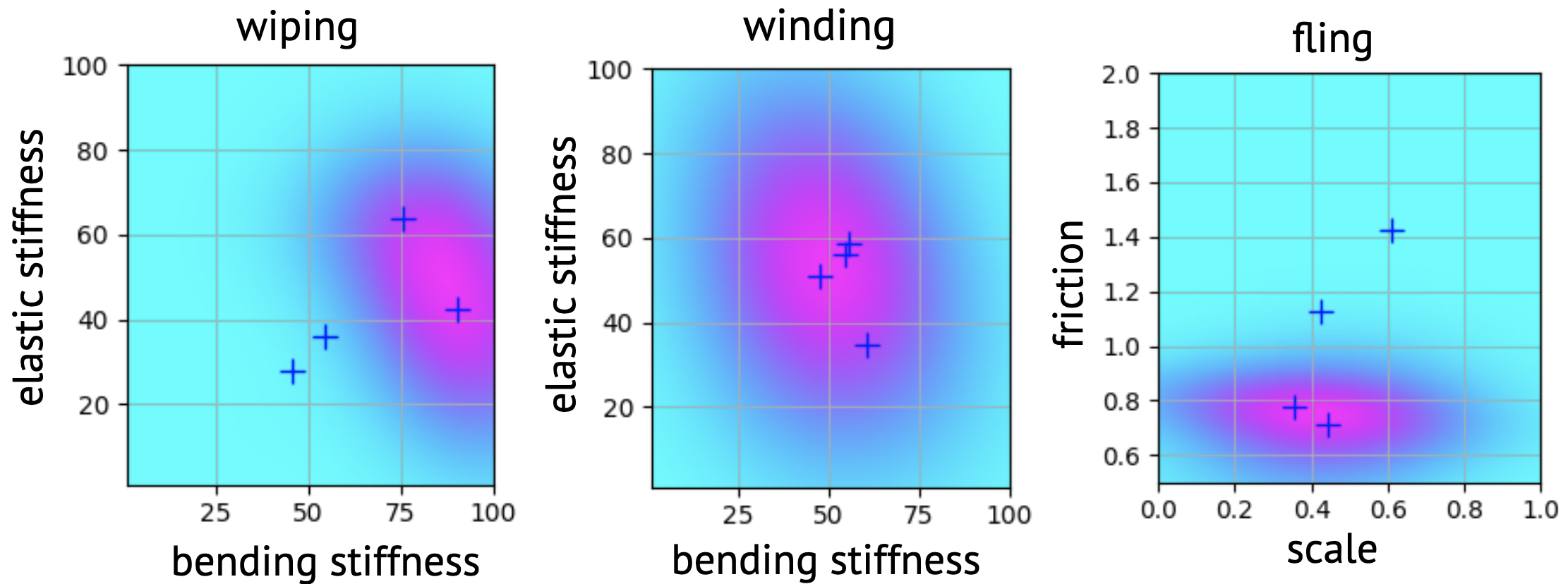}
\vspace{-15px}
\caption{Examples of 2D slices of 4D posteriors found after 15 iterations of \textit{BayesSim-RKHS}. The mixture posterior is comprised of 4 full-covariance Gaussian components, blue crosses show their means. The high-likelihood regions are denoted in magenta (blue crosses outside high-likelihood regions indicate that the mixture component's weight is low). Scale and friction tend to be the easier parameters to estimate, with posteriors becoming peaked. Bending and elastic stiffness are more difficult to infer. The middle plot shows an example where the posterior has shifted only slightly.}
\label{fig:hw_posteriors}
\end{figure}

\begin{figure}[t]
\vspace{10pt}
\centering
\includegraphics[width=0.9\linewidth]{img/legend.jpg}
\includegraphics[width=0.50\linewidth]{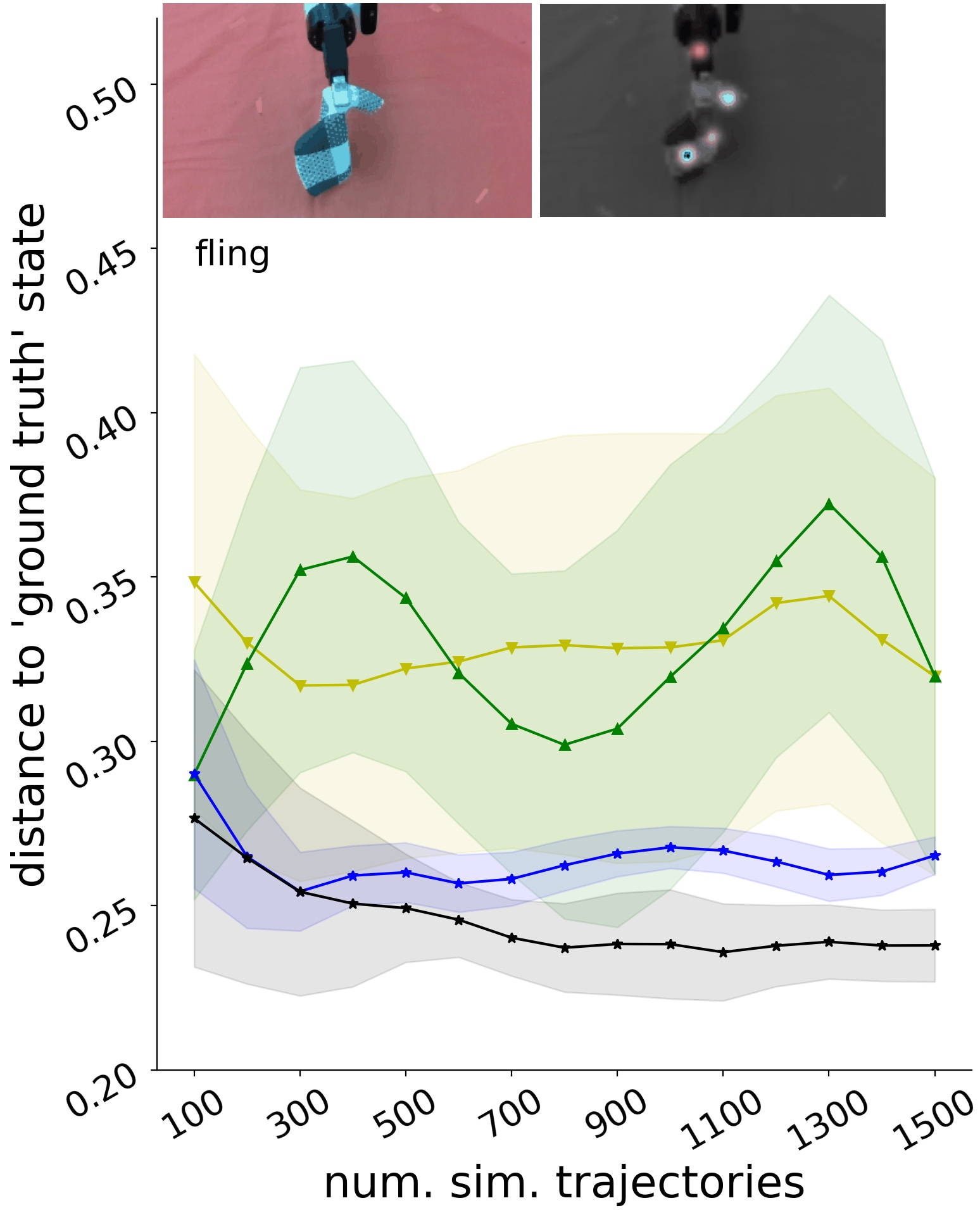}
\includegraphics[width=0.485\linewidth]{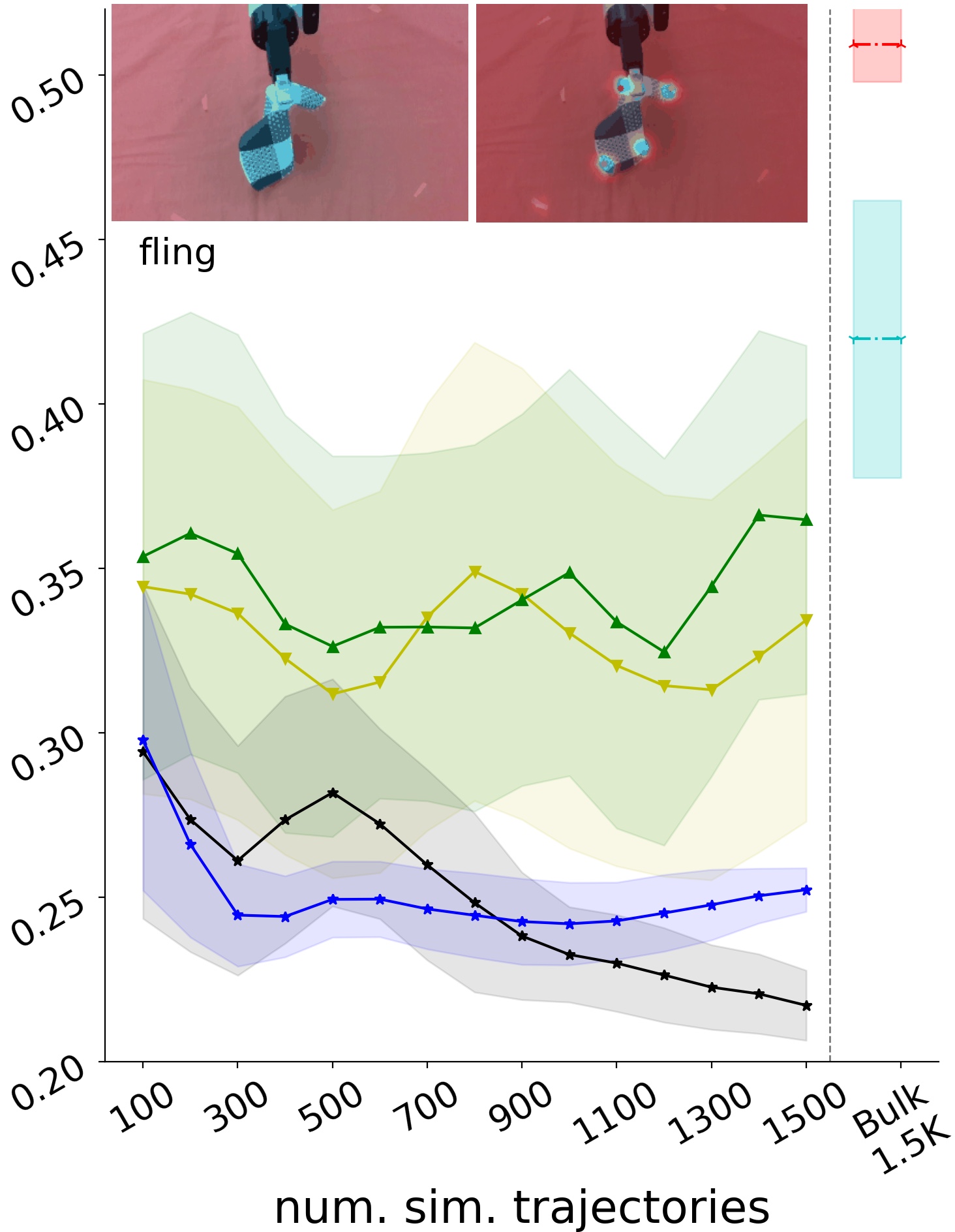}
\vspace{-20px}
\caption{Hardware results for the fling task. Left: using unsupervised keypoints. Right: using supervised keypoints.}
\label{fig:hw_results_fling}
\vspace{-13px}
\end{figure}

Figures~\ref{fig:hw_results_wiping},~\ref{fig:hw_results_winding},~\ref{fig:hw_results_fling} show plots for results on all the scenarios. In these plots, the lines show mean distance to `ground truth' (mean over a set of 30 evaluation trajectories), shaded regions indicate one standard deviation. The top of each plot includes an example visualization of the keypoints. Results for the `bulk' baselines are shown on the right side in each figure. We let these methods use the supervised keypoints, since these are less noisy.
\textit{BayesSim-RKHS} and \textit{BayesZoom-RKHS} outperform all other approaches, showing the largest improvement on the winding task. This result is intuitive, since the keypoints in this scenario move around the length of the rope without settling on consistent locations. Hence, the distributional embedding for keypoints is most useful. The unsupervised keypoints extracted in wiping and fling scenarios are sometimes placed on the robot. This can be resolved by modeling the robot in simulation instead of masking it out in reality; we will address this in future work.

\section{Discussion and Conclusion}

In this work, we introduced the concept of distributional embedding to represent deformable object state. We showed that this idea allows us to conduct Bayesian parameter inference of material properties on noisy real-world data from vision systems. Using random Fourier features approximation enabled this embedding to be computed efficiently.

Previous works, including existing BayesSim variants, explored using RFF features to transform the inputs to the neural networks before training. Computer vision works, such as~\cite{tancik2020fourier}, also employed RFFs and showed favorable results in some cases, for example -- improving recovery of high-frequency signals in images. In contrast to previous results for direct application of RFFs, we show that simply using RFF features yields \mbox{\textit{BayesSim-MDRFF}} method, which is unable to produce informative parameter posteriors for deformable objects.

Our approach benefits from the favorable theoretical properties of RFFs, but applies them in a different context: to embed distributions representing the state of a deformable object in an RKHS. We specifically address the challenges of handling approximate and noisy representations that frequently arise when dealing with deformable objects.
Furthermore, our approach aims to enable modularity and data efficiency when applying Bayesian parameter inference methods to the challenges of learning from data with deformables. 
We demonstrate how to successfully use existing state representation learning methods, despite the lack of consistency and challenges with non-identifiability in these representations. Instead of requiring large-scale data collection, we focus on utilizing the data efficiently. The keypoint extraction models are trained from $\approx\!\!1$ minute of real data, hence output approximate and noisy results. Our method allows to interpret the output of these methods as samples from an underlying state distribution and create embeddings of these distributions with minimal loss of information. The RKHS-net layer we propose offers a fully automated way to construct such embeddings, without the need for hyperparameter tuning, since all variables and parameters are learned from data via gradient descent.

\bibliographystyle{IEEEtran}
\bibliography{references}

\end{document}